\documentclass{article}
\usepackage{spconf,amsmath,epsfig}
\usepackage{ctable}
\usepackage{multirow}
\usepackage{subcaption}

\title{Is Pretraining Necessary for Hyperspectral Image Classification?}
\name{Hyungtae Lee$^{\star\dagger}$~~~~~~~~~~~~~~~~~~~~~~~~Sungmin Eum$^{\star\dagger}$~~~~~~~~~~~~~~~~~~~~~~~~Heesung Kwon$^{\dagger}$}
\address{$^{\star}$Booz Allen Hamilton Inc.~~~~~~~~~~~~~~~~~~~~~~~~$^{\dagger}$US Army Research Laboratory}
\begin{document}
%\ninept
%
\maketitle
\begin{abstract}
We address two questions for training a convolutional neural network (CNN) for hyperspectral image classification: i) is it possible to build a pre-trained network? and ii) is the pre-training effective in furthering the performance? To answer the first question, we have devised an approach that pre-trains a network on multiple source datasets that differ in their hyperspectral characteristics and fine-tunes on a target dataset. This approach effectively resolves the architectural issue that arises when transferring meaningful information between the source and the target networks. To answer the second question, we carried out several ablation experiments. Based on the experimental results, a network trained from scratch performs as good as a network fine-tuned from a pre-trained network. However, we observed that pre-training the network has its own advantage in achieving better performances when deeper networks are required.
\end{abstract}
\section{Introduction}
\label{sec:intro}

In many classification tasks, convolutional neural network (CNN) has been showing a series of innovative performances. However, when only given a small-sized target dataset, it is difficult to avoid the overfitting issue due to a large number of parameters that need to be optimized in a CNN. One widely known approach to go around this issue is to fine-tune the network from the first few layers of a pre-trained network which was trained on a large-scale dataset \cite{RGirshickTPAMI2016}. However, this approach can be applied effectively when the source and the target datasets share equivalent spectral characteristics (e.g., RGB to RGB). When these characteritics do not match between the source and target domains as with the hyperspectral datasets acquired with different sensors, it becomes a challenge in transferring the information effectively, thus posing a problem in constructing a pre-trained model.

In this paper, we have devised an approach which allows the pre-training of a network on multiple hyperspectral ``source'' datasets and uses the pre-trained network to be fine-tuned on a ``target'' dataset. Our fine-tuning process is heavily motivated by the shared layers in the Cross-Domain CNN \cite{HLeeIGARSS2018}. In \cite{HLeeIGARSS2018}, the shared portion is used to train effectively on multiple hyperspectral datasets with different characteristics. In our approach, we exploit this set of shared layers to be used as the ``pre-trained'' portion for the fine-tuning process, and show that it is indeed possible to build a pre-trained network for hyperspectral image classification.

Although pre-trained networks have been used without any doubt in the past several years, researchers recently have begun a comprehensive analyses on the effectiveness of pre-training. Mahajan et al.~\cite{DMahajanECCV2018} analyze the effect of the pre-trained network when increasing the pre-training dataset size. To significantly increase the dataset size, \cite{DMahajanECCV2018} collect social media images and adopt weakly supervised strategy due to a lack of labels of these images. When dataset scale was extremely enlarged, the classification performance was proportionally increased. He et al.~\cite{KHeArxiv2018} questioned whether using a pre-trained network actually increases the classification performance. According to \cite{KHeArxiv2018}, a randomly initialized network provides compatible accuracy to the network fine-tuned from a pre-trained network as long as it is trained with extremely large amount of training time. Based on this observation, they conclude that using pre-trained networks trained on large datasets is not requirement in achieving high accuracy.

Following the recent trends in carrying out comprehensive studies on using pre-trained models, we address relevant questions for hyperspectral image domain: 
\begin{enumerate}
    \item \textit{Is pre-trained network necessary for hyperspectral image classification?}
    \item \textit{Does a larger source dataset improve accuracy?}
    \item \textit{Is pre-training effective when target and source datasets are obtained from different sensors?}
    \item \textit{Does introducing more variety in the source datasets for pre-training increase the performance?}
\end{enumerate}
%To answer these questions properly, we carried out several ablation experiments.

\section{Method}
\label{sec:method}

\noindent{\bf Backbone CNN Architecture.} For the analysis, we used 9-layer fully convolutional network architecture introduced in~\cite{HLeeIGARSS2016,HLeeTIP2017}. The anterior part of the network is composed of a multi-scale filter bank and one convolutional layer ($C_2$). Two residual modules (res\{$C_3$, $C_4$\}, res\{$C_5$, $C_6$\}), and three convolutional layers ($C_7$, $C_8$, $C_9$) are appended after that. Multi-scale filter bank consists of three types of convolutional filters whose dimensions are $1\times1$ ($C_{1\times1}$), $3\times3$ ($C_{3\times3}$), and $5\times5$ ($C_{5\times5}$). The residual modules are connected via skip connection, where each module consists of two convolutional layers. At the end of $C_7$ and $C_8$, dropout layers are attached during training. Batch normalization and ReLU (Rectified Linear Unit) are appended at the end of all convolutional layers except the last layer. Each layer consists of 128 filters.\\

\begin{figure}
  \centering
  \includegraphics[width=0.9\linewidth,trim=5mm 5mm 5mm 5mm,clip]{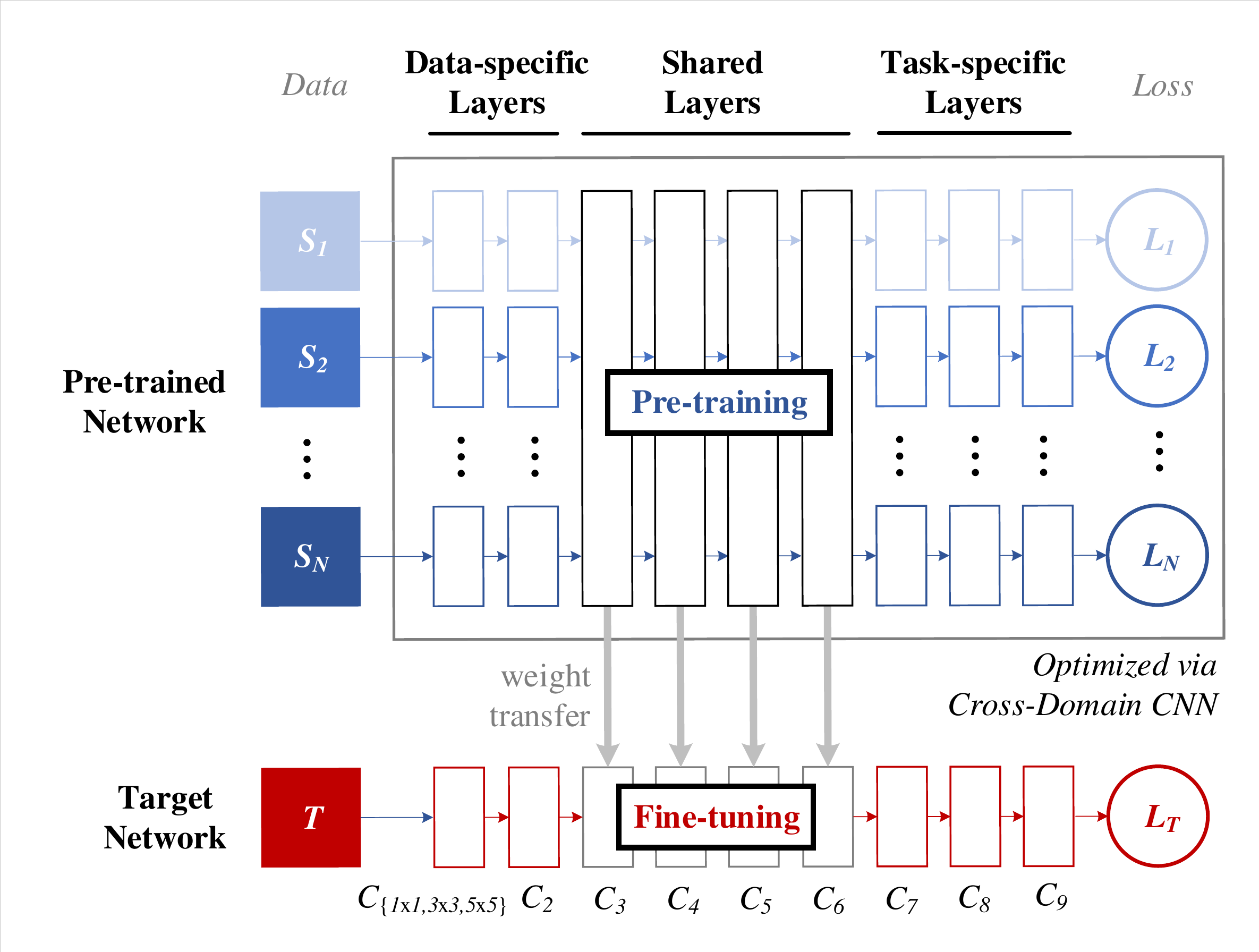}
   \vspace{-0.3cm}
  \captionof{figure}{{\small {\bf Pre-training and fine-tuning.} A pre-training network is trained with $N$ source datasets ($S_1$, $S_2$, $\cdots$, $S_N$) by minimizing $N$ losses ($L_1$, $L_2$, $\cdots$, $L_N$) via Cross-Domain CNN. When a target network is trained, middle shared layers transferred from the pre-trained network are fine-tuned to a target dataset $T$ by minimizing loss $L_T$.}}
  \label{fig:pretraining}
\end{figure}

\noindent{\bf Cross-Domain CNN.} Lee et al.~\cite{HLeeIGARSS2018} introduced Cross-Domain CNN which simultaneously trains multiple networks for classifying multiple hyperspectral image domains. In our implementation, all the networks have the same CNN backbone except the last layer for each network because the dimension has to match the number of categories for the corresponding domain. Weights of two residual modules are shared across all the networks so as to guide the weights to be tuned to a large combined dataset, rather than to fit towards each independent dataset. This sharing mechanism increases the classification accuracy over all the domains.\\

\noindent{\bf Pre-training and Fine-tuning.} We can observe from \cite{HLeeIGARSS2018} that the middle portion of a hyperspectral CNN can serve as a highly beneficial portion which carries common representation across various datasets with different spectral characteristics. In constructing a pre-trained hyperspectral CNN for all the following experiments, we train a Cross-Domain CNN whenever more than one dataset (``source'' dataset) are involved in training. In making use of a pre-trained network for domain adaptation towards a ``target'' dataset, we fine-tune the middle portion (i.e., two residual modules) while other network parameters are learned from scratch. This pre-training and fine-tuning process is illustrated in Figure~\ref{fig:pretraining}.

\section{Experiments}
\label{sec:exp}

\subsection{Settings}
\label{ssec:settings}

\noindent{\bf Dataset.} Six hyperspectral image datasets shown in Table~\ref{tab:dataset} are used for the experiments. Indian Pines dataset is used as the target domain while various combinations of the remaining five datasets are used as source domains. When a network is trained on a target domain, 200 pixels randomly selected from each category are used for training and all the remaining pixels are used for testing. To train each pre-trained network, all the pixels from source datasets are used.\\

\begin{table}
\setlength{\tabcolsep}{1.0pt}
\renewcommand{\arraystretch}{1.1}
{\small
\begin{tabular}{c|c|cccc}
\specialrule{.15em}{.05em}{.05em} 
Domain & Dataset & \# Class & \# Data & Sensr & Bands \\\specialrule{.15em}{.05em}{.05em}
Target & Indian Pines ({\bf I}) & 8 & 8504 & {\bf A} & 200\\\hline
\multirow{5}{*}{Source} & Salinas ({\bf S}) & 17 & 54129 & {\bf A} & 204 \\
& Pavia Centre ({\bf PC}) & 10 & 148152 & {\bf R} & 102 \\
& Pavia University ({\bf PU}) & 10 & 42776 & {\bf R} & 103 \\
& Kennedy Space Center ({\bf KSC}) & 14 & 5211 & {\bf A} & 176 \\
& Botswana ({\bf B}) & 15 & 3248 & {\bf H} & 145
\\\specialrule{.15em}{.05em}{.05em} 
\end{tabular}
}
\vspace{-0.3cm}
\caption{{\small {\bf Hyperspectral datasets.} 
The notation for each dataset is shown next to dataset name. We use these notations in this paper. For each dataset, the number of frequency bands of the dataset acquired by the same sensor may be different because some bands were deleted according to the dataset task. Sensor notations: {\bf A}: AVIRIS, {\bf R}: ROSIS, {\bf H}: Hyperion}}
\label{tab:dataset}
\end{table}

\noindent{\bf Optimization.} We have used stochastic gradient descent to train the networks with the batch size of 128 examples, momentum of 0.9, gamma of 0.1, and weight decay of 0.0005. To avoid overfitting while training, data is augmented 8 times by mirroring each example horizontally, vertically, and diagonally. When the network is trained from scratch, $C_{\{1\times 1,3\times 3,5\times 5\}}$, $C_2$, and $C_9$ are initialized according to a Gaussian distribution with mean of zero and standard deviation of 0.01. The remaining layers are initialized with standard deviation of 0.005. Base learning rate is set as 0.001 which is divided by 10 for every step size. Table~\ref{tab:Iteration} shows different step sizes and iterations for different training sets.

When learning the network on multiple datasets, the base learning rate of the shared layers are multiplied by 1/$N$, where $N$ is the number of the datasets involved in training. This is because the shared layers are affected by multiple back-propagations from multiple losses.

When using all five source datasets ({\bf P5}) for training, we should consider a potential issue caused by the imbalanced dataset sizes. Since the Pavia Center dataset has much more data than the others, it requires more iterations than the others. To cope with this issue, we adopt a two-step optimization strategy introduced in \cite{HLeeICIP2017,HLeeArxiv2018DODCNN}. Under this scheme, the network is initially trained on only the largest dataset (Step I (PC)) and then is updated using the whole dataset for multi-task learning (Step II).

\begin{table}
\setlength{\tabcolsep}{1.3pt}
\renewcommand{\arraystretch}{1.1}
{\small
\begin{tabular}{c|c|ccccc}
\specialrule{.15em}{.05em}{.05em} 
Dataset & {\bf I} & \multicolumn{2}{c}{{\bf P5}} & {\bf P4} &
{\bf P3} & {\bf P2} \\\specialrule{.15em}{.05em}{.05em}
\multirow{2}{*}{SS/Iter.} & \multirow{2}{*}{4K/5K} & Step I (PC) & Step II & \multirow{2}{*}{20K/50K}& \multirow{2}{*}{10K/25K} & \multirow{2}{*}{10K/25K}\\
& & 40K/100K & 20K/50K & & &
\\\specialrule{.15em}{.05em}{.05em} 
\end{tabular}
}
\vspace{-0.3cm}
\caption{{\small {\bf Step size and iteration} for different training sets. Notations of pre-training datasets: {\bf P5}: {\bf S}+{\bf PC}+{\bf PU}+{\bf KSC}+{\bf B}, {\bf P4}: {\bf S}+{\bf PU}+{\bf KSC}+{\bf B}, {\bf P3}: {\bf PU}+{\bf KSC}+{\bf B}, and {\bf P2}: {\bf KSC}+{\bf B}}}
\label{tab:Iteration}
\end{table}

\subsection{Ablation Studies}
\label{ssec:ablation}
Note that for training a network on a target domain, 200 pixels are randomly selected from each category for training while the rest is used for testing. We use the same set consistently throughout all the experiments in the ablation studies. All accuracies reported in the ablation studies are calculated with the target dataset (Indian Pines).\\

\begin{figure*}
\centering
\begin{minipage}{.33\textwidth}
  \vspace{-0.0cm}
  \centering
  \includegraphics[width=0.85\linewidth,trim=5mm 0mm 5mm 0mm,clip]{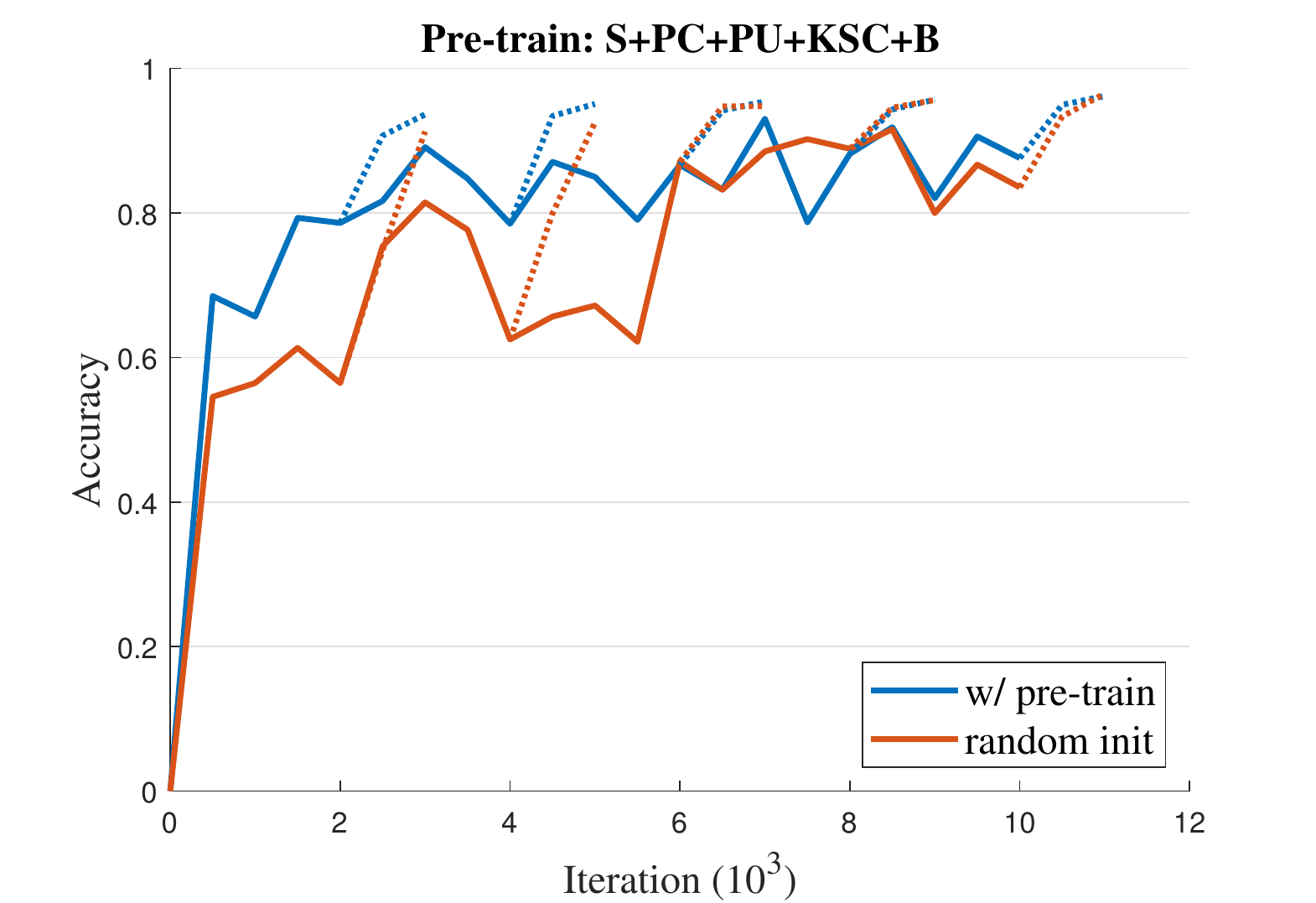}
   \vspace{-0.3cm}
  \captionof{figure}{{\small {\bf Accuracy changes} as the training process evolves. We use a different training schedule by changing the step size. The dotted line shows the change in performance after the step size.}}
  \label{fig:accuracy}
\end{minipage}~~~~~
\begin{minipage}{.33\textwidth}
  \centering
  \includegraphics[width=0.85\linewidth,trim=5mm 0mm 5mm 0mm,clip]{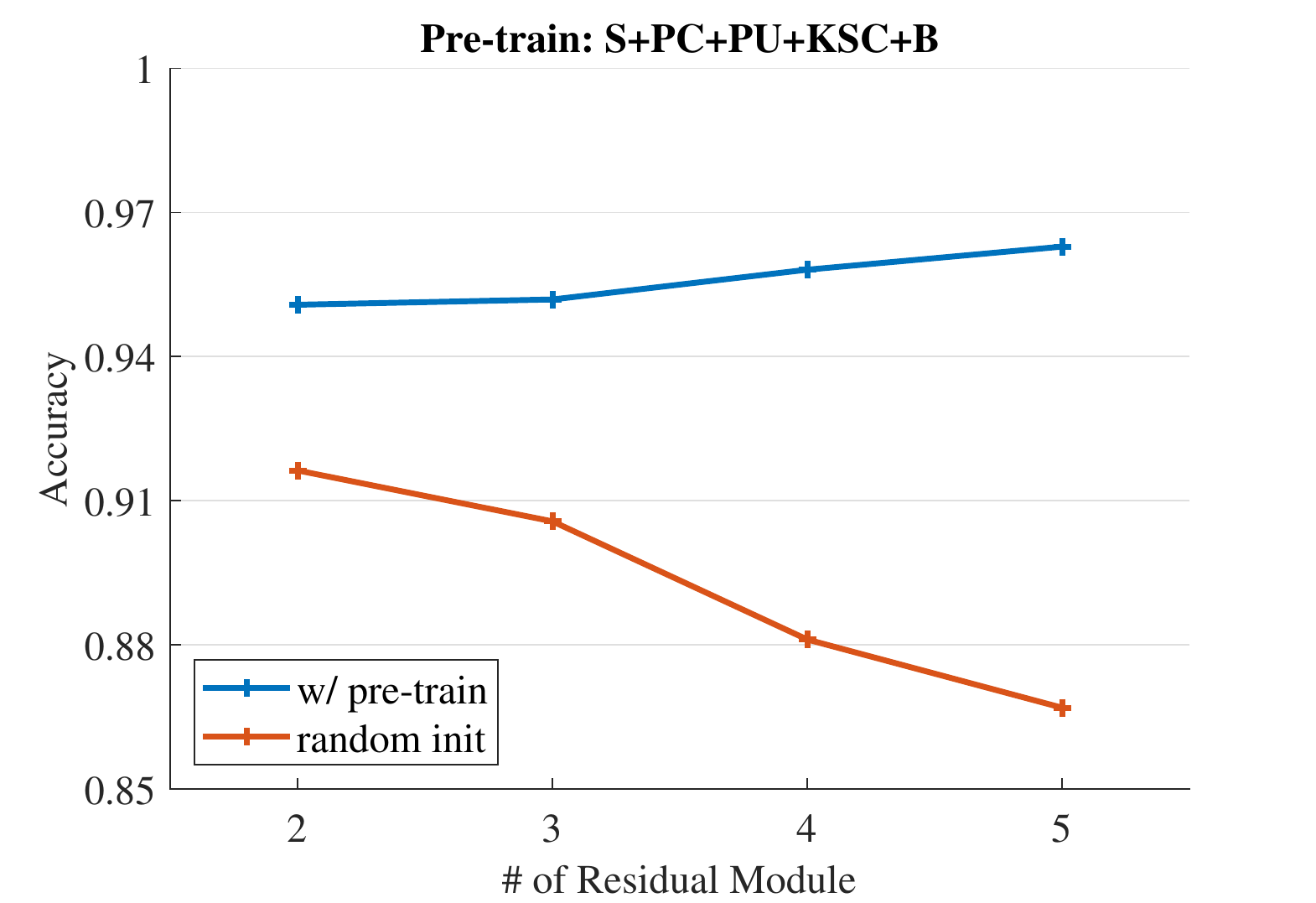}
   \vspace{-0.3cm}
  \captionof{figure}{{\small {\bf Accuracy w.r.t. network depth.} The depth of the network is increased by adding the residual modules. For example, the network that employs 4 residual modules becomes a 13-layer CNN.}}
  \label{fig:going_deeper}
\end{minipage}~~~~~
\begin{minipage}{.3\textwidth}
  \vspace{-0.65cm}
  \centering
  \includegraphics[width=0.85\linewidth,trim=15mm 10mm 10mm 15mm,clip]{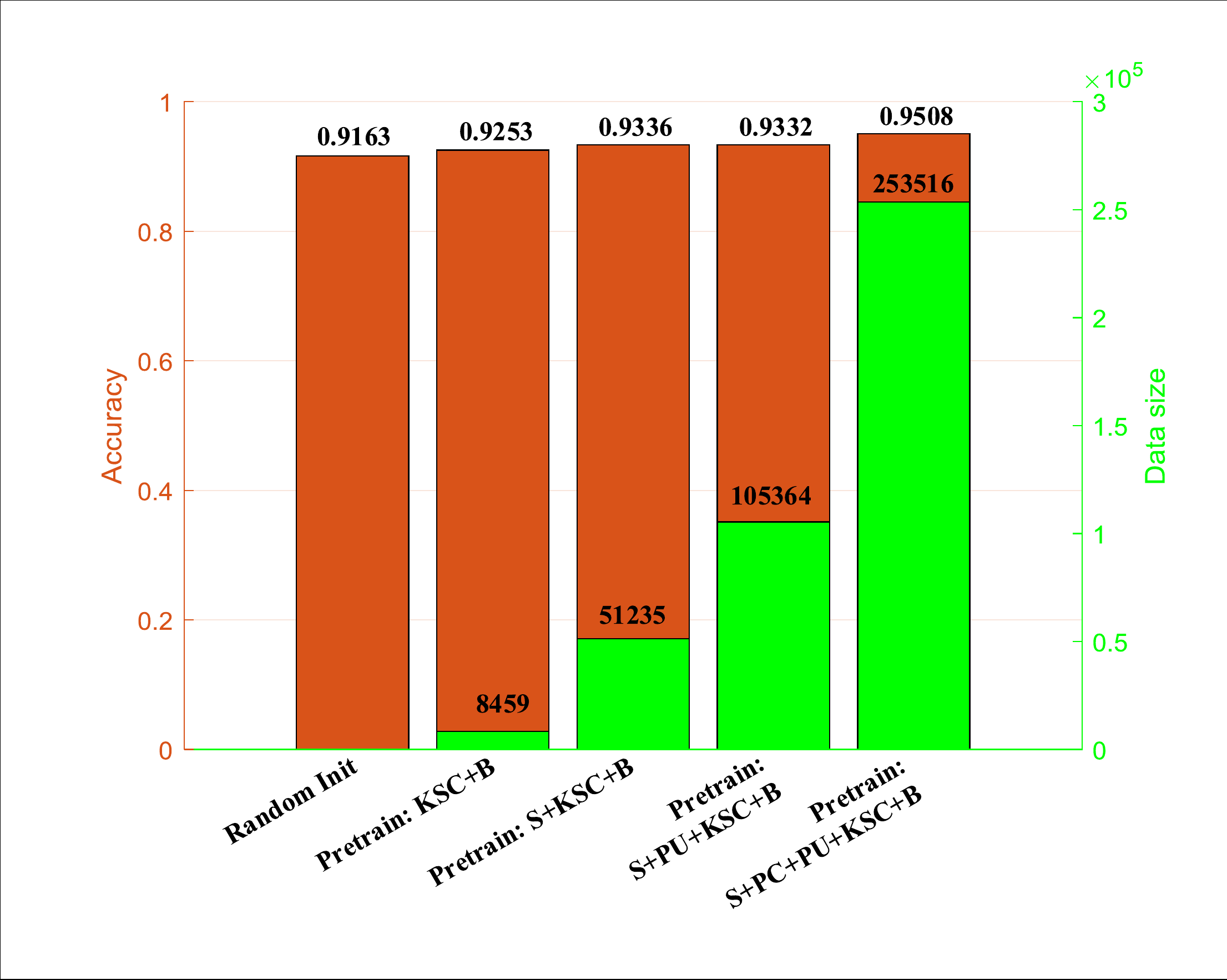}
  \vspace{-0.3cm}
  \captionof{figure}{{\small {\bf Accuracy and dataset size} of randomly initialized network and four combinations of training sets.}}
  \label{fig:pretrain_domain_size}
\end{minipage}
\end{figure*}

\noindent{\bf Training from Scratch vs. Using Pre-training.} Figure~\ref{fig:accuracy} compares how the classification accuracy progresses when trained from scratch (random initialization of weights) and from a pre-trained network. Several different training schedules have been tested and the resulting accuracy is listed in Table~\ref{tab:accuracy_evolution}. Training schedules are varied by changing the maximum iteration and the step size. The base learning rate is 0.001 and is dropped to 0.0001 at the step size. In this experiment, we have used all five source datasets to train a pre-trained network. 

From Figure~\ref{fig:accuracy} and Table~\ref{tab:accuracy_evolution}, we observe that the network using pre-training converges faster than its counterpart. However, when trained with sufficient amount of iterations, the randomly initialized network eventually reaches an accuracy comparable to the one trained on a pre-trained network (0.9652 vs. 0.9618). 

For the remaining experiments, we used accuracy of the network trained with step size/iterations of 4K/5K.\\

\begin{table}
\setlength{\tabcolsep}{5.8pt}
\renewcommand{\arraystretch}{1.1}
{\small
\begin{tabular}{c|ccccc}
\specialrule{.15em}{.05em}{.05em}
SS/Iter. & 2K/3K & 4K/5K & 6K/7K & 8K/9K & 10K/11K \\\specialrule{.15em}{.05em}{.05em}
w/ pre-train & 0.9363 & 0.9508 & 0.9545 & 0.9567 & 0.9618 \\
random init & 0.9127 & 0.9253 & 0.9477 & 0.9563 & 0.9652
\\\specialrule{.15em}{.05em}{.05em} 
\end{tabular}
}
\vspace{-0.3cm}
\caption{{\small {\bf Classification accuracy} when varying the training schedules. SS: Step size.}}
\label{tab:accuracy_evolution}
\end{table}

\noindent{\bf Going Deeper with Residual Modules.} If a large-scale data is available for training, a network with a large number of parameters can be trained with less probability of hitting the overfitting issue. Based on this notion, a pre-trained network which was trained with a large dataset may allow a better chance of training a deeper or wider network which contains more weights. To validate, we compare how the two cases (random init. vs. pre-trained) behave when the depth of the networks is increased with additional residual modules. The accuracy comparison is depicted in Figure~\ref{fig:going_deeper}. In this experiment, we have used all five source data sets to train a pre-trained network.

%\textcolor{red}{[remove this paragraph if the revised paragraph looks OK]}Therefore, using a pre-trained network with a large dataset may allow to train a deeper or wider network that uses more weights. To confirm this effect caused by using the pre-trained network, we increase the depth of the network by adding more residual modules. Figure~\ref{fig:going_deeper} shows accuracy of different networks with different depths for both using pre-trained network and random initialization. In this experiment, we have used all five source data sets to train a pre-trained network.

While the accuracy of randomly initialized network decreases greatly as the network gets deeper, the pre-trained network rather improves its performance. This supports our argument that fine-tuning from a pre-trained network is beneficial to avoid the overfitting issue when the network gets deeper. In addition, 9-layer CNN using pre-training may be underfitted.\\
%In the case of 9-layer CNN \textcolor{red}{(??)}, the fact that both cases show similar accuracy may indicate that the CNN is underfitted on the target training data set.\\

\noindent{\bf Source Data Size.} In this section, we question how much influence the size of a pre-training dataset has over the classification performance. To answer this question, we generate four pre-training datasets by combining two or more source datasets and use them to train separate networks. Figure~\ref{fig:pretrain_domain_size} shows the classification accuracy and dataset sizes of these four cases. From this figure, we can confirm that the larger the size of the data used for pre-training, the better the performance.\\

\noindent{\bf Sources from the Same Sensor vs. Multiple Sensors.} We verify the effect of pre-training a network on a source dataset acquired by a sensor different from the target data.
%\textcolor{red}{(remove next sentence if previous sentence is OK.)} We verify the effect of fine-tuning on the network pre-trained with the source data obtained by the sensor different from the sensor of the target data. 
In general, using a pre-trained network to fine-tune is effective if the source data which is used for pre-training and the target data shares the same domain.
%\textcolor{red}{(remove next sentence if previous sentence is OK.)}
%In general, it is effective to pre-train images in the same domain with target images. 
In this experiment, we prepare two different pre-training datasets. One is {\bf S}+{\bf KSC} which is acquired using the same sensor used to acquire the target data (AVIRIS sensor) and the other is {\bf PU}+{\bf B} which is obtained by different sensors, ROSIS and the Hyperion sensor. These two datasets contain similar numbers of examples (59K and 46K).
%\textcolor{red}{(remove next sentence if previous two sentences are OK.)}
%In this experiment, we prepare two pre-training sets of {\bf S}+{\bf KSC} obtained from the AVIRIS sensor used for the target data and {\bf PU}+{\bf B} obtained by ROSIS and the Hyperion sensor. These two datasets contain similar numbers of data (59K and 46K).

Table~\ref{tab:sensors} shows the classification accuracy when using these two pre-training sets.
Regardless of the fact that {\bf PU}+{\bf B} is smaller in data size (46024 vs. 59340) and that it does not match the source target data in terms of the acquisition sensor when compared with {\bf S}+{\bf KSC}, its performance is better (0.9286 vs. 0.9217).
%\textcolor{red}{(remove next sentence if previous two sentences are OK.)}
%Note that {\bf PU}+{\bf B} is smaller in data size than {\bf S}+{\bf KSC} (46024 vs 59340), but its performance is better (0.9286 vs 0.9217). 
This observation indicates that our pre-training and fine-tuning strategy is effective regardless of source data characteristics.\\

\begin{table}
\setlength{\tabcolsep}{6.3pt}
\renewcommand{\arraystretch}{1.1}
{\small
\begin{tabular}{cc|ccc|c}
\specialrule{.15em}{.05em}{.05em}
Target & Sensor & Source & Sensor & $\#$ of data & Accuracy \\\specialrule{.15em}{.05em}{.05em}
\multirow{2}{*}{\bf{I}} & \multirow{2}{*}{\bf{A}} & {\bf S}+{\bf KSC} & {\bf A} & 59340 & 0.9217 \\
&&{\bf PU}+{\bf B} & {\bf R} \& {\bf H} & 46024 & 0.9286
\\\specialrule{.15em}{.05em}{.05em} 
\end{tabular}
}
\vspace{-0.3cm}
\caption{{\small {\bf Accuracy comparison w.r.t. source data sensors.} Sensor notations: {\bf A}: AVIRIS, {\bf R}: ROSIS, {\bf H}: Hyperion}}
\label{tab:sensors}
\end{table}

\noindent{\bf Single Source vs. Multiple Sources.} The source training dataset can be either a single dataset or a combination of multiple datasets. For optimizing a network with multiple datasets, we apply Cross-Domain CNN approach.
%\textcolor{red}{(remove next sentence if previous two sentences are OK.)}
%Here, CNN can simultaneously be optimized with multiple datasets via Cross-Domain CNN. To verify the effectiveness of Cross-Domain CNN\textcolor{red}{(Why verify the effectiveness of Cross-Domain??)},
We compare the accuracy of using single source dataset ({\bf PC}) and multiple source dataset ({\bf P4} and {\bf P5}). The results are shown in Table~\ref{tab:single_vs_multiple}. 

When using single source dataset, its accuracy was worse than using {\bf P5} but better than using {\bf P4}. This trend indicates that the accuracy improves proportional to the number of data regardless of whether you are using single or multiple datasets.

\begin{table}
\setlength{\tabcolsep}{14.2pt}
\renewcommand{\arraystretch}{1.1}
{\small
\begin{tabular}{c|c|c}
\specialrule{.15em}{.05em}{.05em}
Source & $\#$ of data & Accuracy \\\specialrule{.15em}{.05em}{.05em}
{\bf P5}: {\bf S}+{\bf PC}+{\bf PU}+{\bf KSC}+{\bf B} & 253516 & 0.9508 \\
{\bf PC} & 148152 & 0.9438 \\
{\bf P4}: {\bf S}+{\bf PU}+{\bf KSC}+{\bf B} & 105364 & 0.9332
\\\specialrule{.15em}{.05em}{.05em} 
\end{tabular}
}
\vspace{-0.3cm}
\caption{{\small {\bf Performance comparison of using single source and multiple sources.}}}
\label{tab:single_vs_multiple}
\end{table}

\section{Discussions}
\label{sec:discuss}

\noindent\textit{\textbf{Is pre-training necessary?}} Yes.

In terms of performance, the network which is trained from scratch shows comparable results when compared with the case where a pre-trained network is used. However, using pre-training with large-scale dataset has other advantages, such as having more potential in gaining the performance when increasing the depth of a network.\\

\noindent\textit{\textbf{Does a larger source dataset improve accuracy?}} Yes.

In our experiment, as the size of the pre-training datasets gets larger, the accuracy achieved via fine-tuning on a pre-trained network improves. This improvement is, in part, acquired as the overfitting issue is reduced with the enlarged source dataset.\\
%seems to resolve the overfitting issue better.This shows that adopting a pre-training dataset can reduce side effects caused by overfitting issue.\textcolor{red}{(missing logical connection between first and second sentences)}\\

\noindent\textit{\textbf{Is a source dataset from different sensors effective?}} Yes.

In our experiment, pre-training was effective regardless of the sensor type by which the source dataset was obtained.\\

\noindent\textit{\textbf{Does introducing more variety in the source datasets for pre-training always increase the performance?}} No.

The experimental results show that the accuracy is more affected by the number of samples in the overall dataset than how many different source datasets are involved in the pre-training.

\bibliographystyle{IEEEbib}
\bibliography{refs}

\end{document}